\documentclass[10pt,twocolumn,letterpaper]{article}

\usepackage{cvpr}
\usepackage{times}
\usepackage{url}
\usepackage{epsfig}
\usepackage{graphicx}
\usepackage{amsmath}
\usepackage{amssymb}
\usepackage{subfigure}
\usepackage{algorithm}
\usepackage{algpseudocode}
\usepackage{algorithmicx}
\usepackage{color}
\usepackage{multirow}

\cvprfinalcopy % *** Uncomment this line for the final submission

\def\cvprPaperID{267} % *** Enter the CVPR Paper ID here

\ifcvprfinal\fi
\begin{document}

%%%%%%%%% TITLE
\title{Cross-Scale Cost Aggregation for Stereo Matching}

\author{
Kang Zhang$^1$, Yuqiang Fang$^2$, Dongbo Min$^3$, Lifeng Sun$^1$, Shiqiang Yang$^1$, Shuicheng Yan$^2$, Qi Tian$^4$\\
$^1$Department of Computer Science, Tsinghua University, Beijing, China\\
$^2$Department of Electrical and Computer Engineering, National University of Singapore, Singapore\\
$^3$Advance Digital Science Center, Singapore\\
$^4$Department of Computer Science, University of Texas at San Antonio, Texas, USA\\
{\tt\small \url{https://github.com/rookiepig/CrossScaleStereo}}
}
% \and
% Yuqiang Fang\\
% \and
% Dongbo Min\\
% \and
% Lifeng Sun\\
% \and
% Shiqiang Yang\\
% \and
% Shuicheng Yan\\
% \and
% Qi Tian\\

\maketitle
%\thispagestyle{empty}

%%%%%%%%% ABSTRACT
\begin{abstract}
Human beings process stereoscopic correspondence across multiple scales. However, this bio-inspiration is ignored by state-of-the-art cost aggregation methods for dense stereo correspondence. In this paper, a generic cross-scale cost aggregation framework is proposed to allow multi-scale interaction in cost aggregation. We firstly reformulate cost aggregation from a unified optimization perspective and show that different cost aggregation methods essentially differ in the choices of similarity kernels. Then, an inter-scale regularizer is introduced into optimization and solving this new optimization problem leads to the proposed framework. Since the regularization term is independent of the similarity kernel, various cost aggregation methods can be integrated into the proposed general framework. We show that the cross-scale framework is important as it effectively and efficiently expands state-of-the-art cost aggregation methods and leads to significant improvements, when evaluated on Middlebury,  KITTI and New Tsukuba datasets.

%Furthermore, even the simple box filtering method becomes very
%powerful when integrated into the cross-scale cost aggregation
%framework.
\end{abstract}

\vspace{-2mm}
%%%%%%%%% BODY TEXT
\section{Introduction} % (fold)
\label{sec:intro}
\vspace{-1mm}

% \begin{figure*}[t]
%   \begin{center}
%     \includegraphics[width=0.9\textwidth]{procedure}
%   \end{center}
%   \caption{Overview of the conventional stereo matching flowchart. In this work we replace cost aggregation with the proposed cross-scale cost aggregation, which leads to boosted performance for state-of-the-art cost aggregation methods.}
%   \vspace{-3mm}
%   \label{fig:proc}
% \end{figure*}

\begin{figure}
    \begin{center}
        \includegraphics[width=0.48\textwidth]{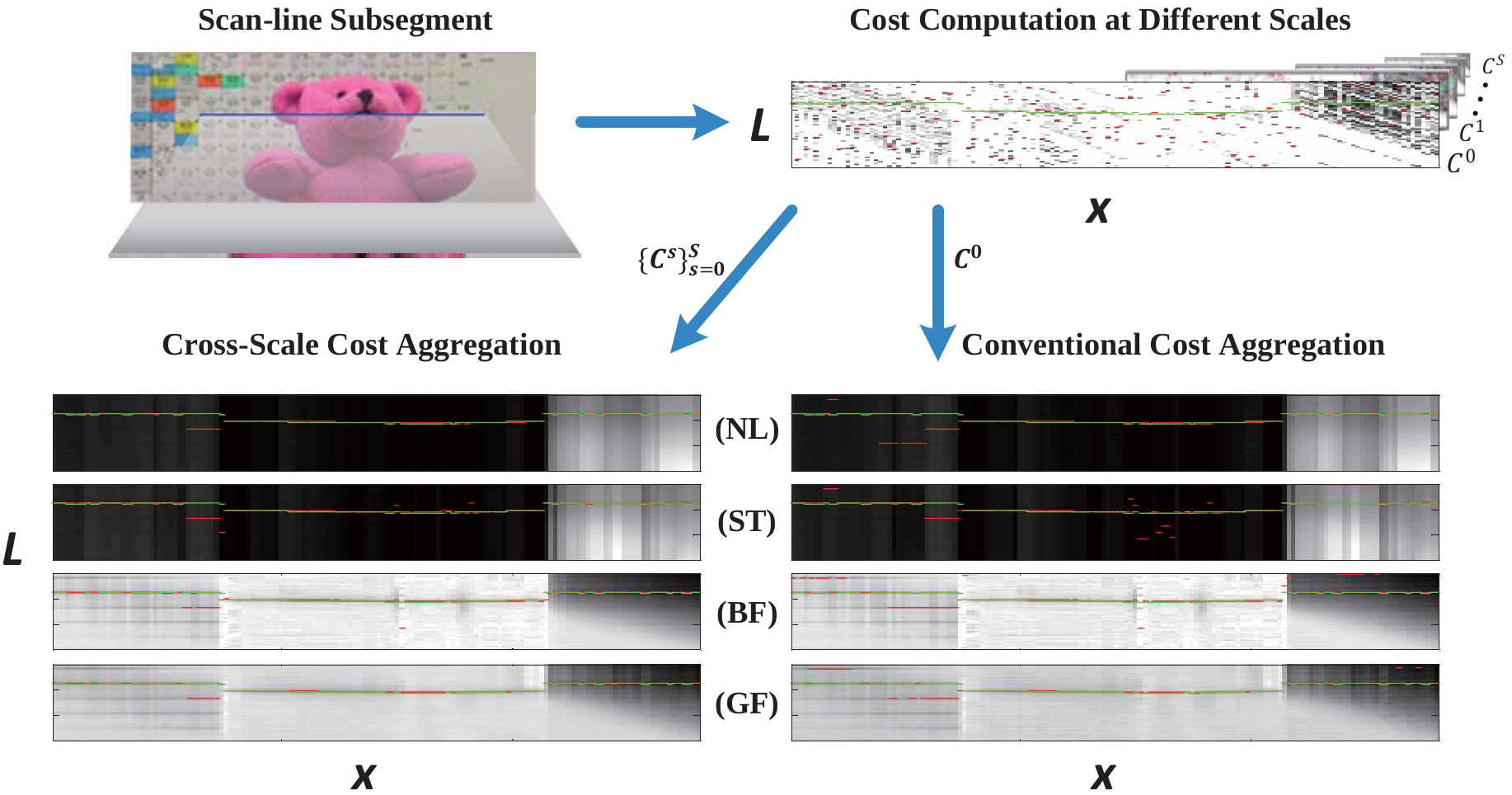}
    \end{center}
    \vspace{-2mm}
       \caption{Cross-Scale Cost Aggregation. \textbf{Top-Left:} enlarged-view of a scan-line subsegment from Middlebury \cite{scharstein_02}  \textit{Teddy} stereo pair; \textbf{Top-Right:} cost volumes ($\{\mathbf{C}^s\}_{s=0}^{S}$) after cost computation at different scales, where the \emph{intensity + gradient} cost function is adopted as in \cite{rhemann_11,Yang2012,XingMei}. Horizontal axis $x$ indicates different pixels along the subsegment, and vertical axis $L$ represents different disparity labels. Red dot indicates disparity generated by current cost volume while green dot is the ground truth; \textbf{Bottom-Right:} cost volumes after applying different cost aggregation methods at the finest scale (from top to bottom: \emph{NL} \cite{Yang2012}, \emph{ST} \cite{XingMei}, \emph{BF} \cite{Yoon2006} and \emph{GF} \cite{rhemann_11}); \textbf{Bottom-Left:} cost volumes after integrating different methods into our cross-scale cost aggregation framework, where cost volumes at different scales are adopted for aggregation. (Best viewed in color.) }
    \vspace{-5mm}
    \label{fig:intro}
\end{figure}

% Introduction
% Para 1:
%   summary stereo algorithms
%   our focus is cost aggregation methods
Dense correspondence between two images is a key problem in computer
vision \cite{liuce_2011}. Adding a constraint that the two images
are a stereo pair of the same scene, the dense correspondence
problem degenerates into the stereo matching problem
\cite{scharstein_02}. A stereo matching algorithm generally takes four steps: \emph{cost computation}, \emph{cost (support) aggregation},
\emph{disparity computation} and \emph{disparity refinement} \cite{scharstein_02}. In cost computation, a 3D cost volume (also known as disparity space image \cite{scharstein_02}) is generated by computing
matching costs for each pixel at all possible disparity levels. In cost
aggregation, the costs are aggregated, enforcing \emph{piecewise
constancy} of disparity, over the support region of each pixel.
Then, disparity for each pixel is computed with local or global
optimization methods and refined by various post-processing methods
in the last two steps respectively. Among these steps, the quality
of cost aggregation has a significant impact on the success of
stereo algorithms. It is a key ingredient for state-of-the-art local
algorithms \cite{Yoon2006,rhemann_11,Yang2012,XingMei} and a primary
building block for some top-performing global algorithms
\cite{Yang2009,Wang2008}. Therefore, in this paper, we mainly concentrate on
\emph{cost aggregation}.

% Introduction
% Para 2:
%   introduce some cost aggregation methods

Most cost aggregation methods can be viewed as joint filtering over
the cost volume \cite{rhemann_11}. Actually, even simple linear
image filters such as box or Gaussian filter can be used for cost
aggregation, but as isotropic diffusion filters, they tend to blur
the depth boundaries \cite{scharstein_02}. Thus, a number of
edge-preserving filters such as bilateral filter \cite{Tomasi1998} %(i.e.\ Yaroslavsky filter \cite{Yaroslavsky1985})
and guided image filter \cite{hekaiming_10} were introduced for cost
aggregation. Yoon and Kweon \cite{Yoon2006} adopted the bilateral
filter into cost aggregation, which generated appealing disparity
maps on the Middlebury dataset \cite{scharstein_02}. However, their
method is computationally expensive, since a large kernel size
(\eg $35 \times 35$) is typically used for the sake of high
disparity accuracy. To address the computational limitation of the
bilateral filter, Rhemann \etal \cite{rhemann_11} %and De-Maeztu \emph{et al.} \cite{de-maeztu_11}
introduced the guided image filter into cost aggregation, whose
computational complexity is independent of the kernel size.
Recently, Yang \cite{Yang2012} proposed a \emph{non-local} cost
aggregation method, which extends the kernel size to the entire
image. By computing a minimum spanning tree (MST) over the image
graph, the non-local cost aggregation can be performed extremely
fast. Mei \etal \cite{XingMei} followed the non-local cost
aggregation idea and showed that by enforcing the disparity
consistency using segment tree instead of MST, better disparity maps
can be achieved than \cite{Yang2012}.

% Introduction
% Para 3:
%   multi-scale aggregation is needed
All these state-of-the-art cost aggregation methods have made great
contributions to stereo vision. A common property of these methods
is that costs are aggregated at the finest scale of the input stereo
images. However, human beings generally process stereoscopic
correspondence across multiple scales \cite{Menz2003,Marr1979,Mallot1996}.
According to \cite{Mallot1996}, information at coarse and fine
scales is processed interactively in the correspondence search of
the human stereo vision system. Thus, from this bio-inspiration, it
is reasonable that costs should be aggregated across multiple scales
rather than the finest scale as done in conventional methods (Figure~\ref{fig:intro}).

% Introduction
% Para 4:
%   introduce our work

In this paper, a general cross-scale cost aggregation framework is
proposed. Firstly, inspired by the formulation of image filters in
\cite{Milanfar2013}, we show that various cost aggregation methods
can be formulated uniformly as weighted least square (WLS)
optimization problems. %Different methods differ in the choices of similarity kernels \cite{Milanfar2013}.
Then, from this unified optimization perspective, by adding a
Generalized Tikhonov regularizer into the WLS optimization
objective, we enforce the consistency of the cost volume among the
neighboring scales, i.e.\ inter-scale consistency. The new
optimization objective with inter-scale regularization is convex and
can be easily and analytically solved. As the intra-scale
consistency of the cost volume is still maintained by conventional 
cost aggregation methods, many of them can be integrated into our
framework to generate more robust cost volume and better disparity
map. Figure~\ref{fig:intro} shows the effect of the proposed framework. Slices of the cost volumes of four representative cost aggregation methods, including the non-local method \cite{Yang2012} (\emph{NL}), the segment tree  method \cite{XingMei} (\emph{ST}), the bilateral filter method  \cite{Yoon2006} (\emph{BF}) and the guided filter method \cite{rhemann_11}  (\emph{GF}), are
visualized. We use red dots to denote disparities generated by local
winner-take-all (WTA) optimization in each cost volume and green
dots to denote ground truth disparities. It can be found that more
robust cost volumes and more accurate disparities are produced by
adopting cross-scale cost aggregation. Extensive experiments on
Middlebury \cite{scharstein_02}, KITTI \cite{Geiger2012} and New
Tsukuba \cite{Peris} datasets  also reveal that better disparity
maps can be obtained using cross-scale cost aggregation. In summary,
the contributions of this paper are three folds: \vspace{-2mm}
\begin{itemize}
  \item A unified WLS formulation of various cost aggregation methods from an optimization perspective.
  \vspace{-2mm}
  \item A novel and effective cross-scale cost aggregation framework.
  \vspace{-2mm}
  \item Quantitative evaluation of representative cost aggregation methods on three datasets.
\end{itemize}
\vspace{-2mm}

% Introduction
% Para 6:
%   organization of the whole paper
The remainder of this paper is organized as follows. In
Section~\ref{sec:rw}, we summarize the related work. The WLS 
formulation for cost aggregation is given in Section~\ref{sec:tso}.
Our inter-scale regularization is described in
Section~\ref{sec:isr}. Then we detail the implementation of our framework in Section~\ref{sec:impl}. Finally experimental results and analyses
are presented in Section~\ref{sec:exp} and the conclusive remarks are
made in Section~\ref{sec:con}.

% \begin{figure*}[tb]
%     \begin{center}
%         \includegraphics[width=0.8\textwidth]{intro}
%     \end{center}
%        \caption{Cross-Scale Cost Aggregation. \textbf{Top-Left:} enlarged-view of a scan-line subsegment from Middlebury \textit{Teddy} stereo pair; \textbf{Top-Right:} cost volumes ($\{\mathbf{C}^s\}_{s=0}^{S}$) after cost computation at different scales, where the \emph{intensity + gradient} cost function is adopted as in \cite{rhemann_11,Yang2012,XingMei}. Horizontal axis $x$ indicates different pixels along the subsegment, and vertical axis $L$ represents different disparity labels. Red dot indicates disparity generated by current cost volume while green dot is the ground truth; \textbf{Bottom-Right:} cost volumes after applying different cost aggregation methods in the finest scale (from top to bottom: \emph{NL} \cite{Yang2012}, \emph{ST} \cite{XingMei}, \emph{GF} \cite{rhemann_11} and \emph{BF} \cite{Yoon2006} ); \textbf{Bottom-Left:} cost volumes after integrating different methods into our cross-scale cost aggregation framework, where cost volumes at different scales are adopted for aggregation. (Best viewed in color.) }
%     \vspace{-5mm}
%     \label{fig:intro}
% \end{figure*}

% section intro (end)

\vspace{-1mm}
\section{Related Work} % (fold)
\label{sec:rw}
\vspace{-1mm}
% Related Work
% Para: 1
%   mainly review multi-scale methods

%In this section, we will review the related work and show the originality of the proposed approach.
Recent surveys \cite{Hosni2013,tombari_08} give sufficient comparison and %tombari_08
analysis for various cost aggregation methods. We refer readers to
these surveys to get an overview of different cost aggregation
methods and we will focus on stereo matching methods involving
multi-scale information, which are very relevant to our idea but
have substantial differences.

Early researchers of stereo vision adopted the coarse-to-fine (CTF)
strategy for stereo matching \cite{Marr1979}. Disparity of a coarse
resolution was assigned firstly, and coarser disparity was used to
reduce the search space for calculating finer disparity. This CTF
(hierarchical) strategy has been widely used in global stereo
methods such as dynamic programming \cite{VanMeerbergen2002},
semi-global matching \cite{Simon2012}, and belief propagation
\cite{Felzenszwalba,Yang2009} %simulated annealing \cite{Chang1990} and partial differential equation (PDE) based approach \cite{Alvarez2002}
for the purpose of accelerating convergence and avoiding unexpected
local minima. Not only global methods but also local methods adopt
the CTF strategy. Unlike global stereo methods, the main purpose of
adopting the CTF strategy in local stereo methods is to reduce the
search space \cite{Pollefeys2003,Yi-Hung2011,Hu2013} or take the
advantage of multi-scale related image representations
\cite{Sizintsev2008,Tang2011}. While, there is one exception in %Magarey1998
local CTF approaches. Min and Sohn \cite{Min2008} modeled the cost
aggregation by anisotropic diffusion and solved the proposed variational model efficiently by the multi-scale approach. The motivation of their model is to denoise the cost volume which is very similar with us, but our
method enforces the inter-scale consistency of cost volumes by
regularization.

% Magarey \textit{et al.} \cite{Magarey1998} adopted the coarse-to-fine framework, based on the Complex Discrete Wavelet Transform (CDWT). The CDWT feature space efficiently provides fractionally accurate matching results. Yang and Pollefeys \cite{Pollefeys2003} proposed a multi-resolution approach to combine the sum-of-square (SSD) cost for windows of different sizes, which mimics a weighted correlation kernel. Sizintsev \cite{Sizintsev2008} proposed to perform CTF stereo matching  in a generalization of the Laplacian pyramid to resolve the problem of poor recovery of thin structures - a common drawback of CTF approach \cite{sizintsev2006}. Tang \textit{et al.} \cite{Tang2011} proposed a robust multiscale stereo matching algorithm to handle fundus images with radiometric differences. They invented the multi-scale pixel feature vector and performed matching in the neighboring scales to generate a disparity map at each scale. Jen \textit{et al.} \cite{Yi-Hung2011} introduced an adaptive scale selection mechanism by convolving the surface prior image with a Laplacian of Gaussian kernel. The scale selection results helped to determine the starting scale level for CTF approach.

Overall, most CTF approaches share a similar property. They
explicitly or implicitly model the disparity evolution process in
the scale space \cite{Tang2011}, i.e.\ \emph{disparity consistency}
across multiple scales. Different from previous CTF methods, our
method models the evolution of the cost volume in the scale space,
i.e.\ \emph{cost volume consistency} across multiple scales. From
optimization perspective, CTF approaches narrow down the solution
space, while our method does not alter the solution space but adds
inter-scale regularization into the optimization objective. Thus,
incorporating multi-scale prior by regularization is the
originality of our approach. Another point worth mentioning is that
local CTF approaches perform no better than state-of-the-art cost
aggregation methods \cite{Hu2013,Yi-Hung2011}, while our method can
significantly improve those cost aggregation methods
\cite{rhemann_11,Yang2012,XingMei}.

%section rw (end)
\vspace{-1mm}
\section{Cost Aggregation as Optimization} % (fold)
\label{sec:tso}
\vspace{-1mm}

\begin{figure}
  \begin{center}
    \includegraphics[width=0.45\textwidth]{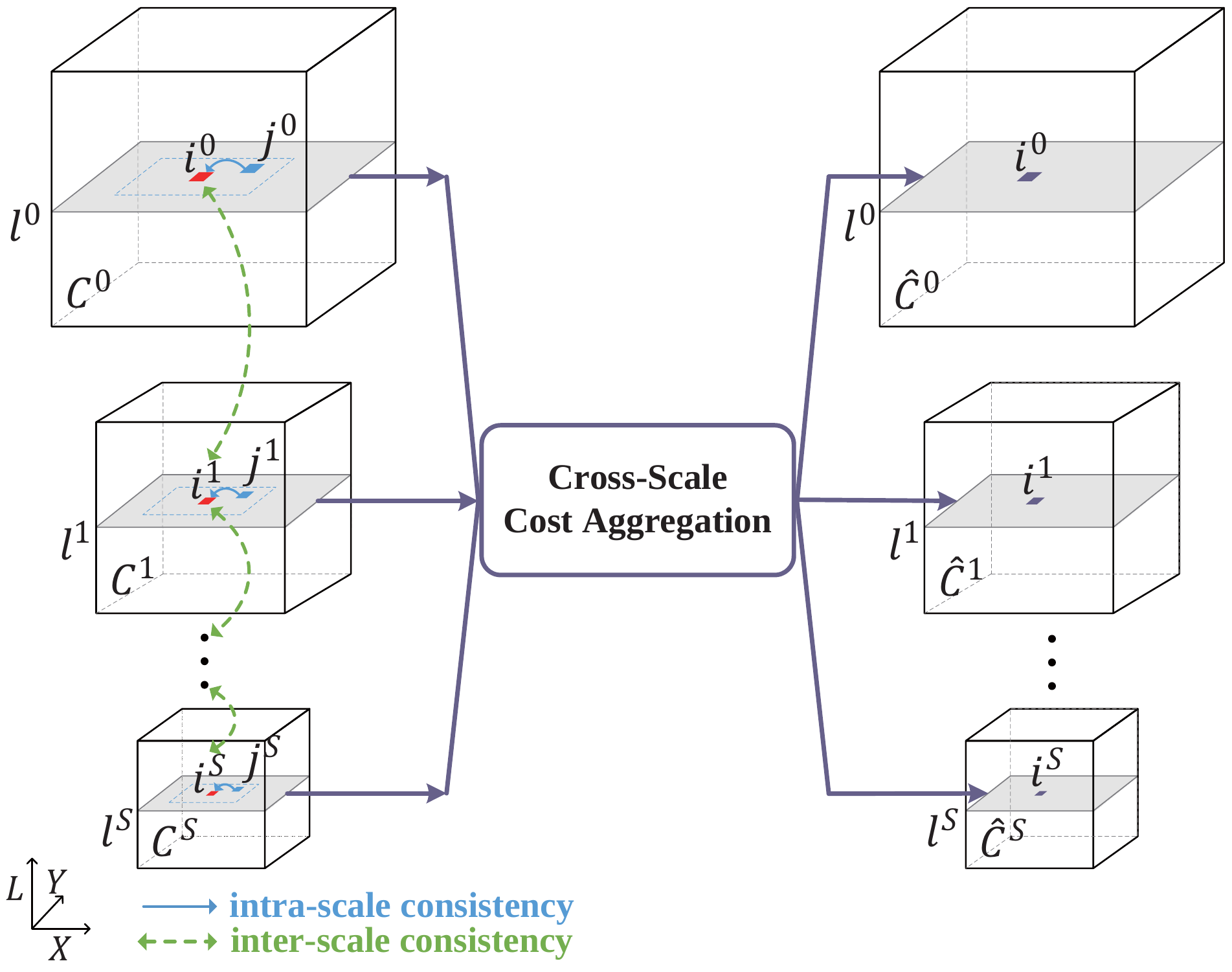}
  \end{center}
  \vspace{-2mm}
  \caption{The flowchart of cross-scale cost aggregation: $\{\mathbf{\hat{C}}^s\}_{s=0}^{S}$
  is obtained by utilizing a set of input cost volumes, $\{\mathbf{C}^s\}_{s=0}^{S}$, together.
  Corresponding variables $\{i^s\}_{s=0}^{S}, \{j^s\}_{s=0}^{S}$ and $\{l^s\}_{s=0}^{S}$ are
  visualized. The blue arrow represents an \emph{intra-scale} consistency (commonly used in the conventional cost aggregation approaches),
  while the green dash arrow denotes an \emph{inter-scale}
  consistency. (Best viewed in color.)}
  \vspace{-3mm}
  \label{fig:csca}
\end{figure}

% Para 1:
%   briefly introduce this section

In this section, we show that the cost aggregation can
be formulated as a weighted least square optimization problem. Under this formulation, different choices of similarity kernels \cite{Milanfar2013} in the optimization objective lead to different cost aggregation methods.

% Para 2:
%   model cost computation

Firstly, the cost computation step is formulated as a
function $f \colon \mathbb{R}^{W \times H \times 3} \times
\mathbb{R}^{ W \times H \times 3 } \mapsto \mathbb{R}^{W \times H
\times L}$, where $W$, $H$ are the width and height of
input images, $3$ represents color channels and $L$ denotes the number of disparity levels. Thus, for a stereo color pair: $\mathbf{I}, \mathbf{I'} \in \mathbb{R}^{ W \times H \times 3 }$, by applying cost
computation:
\begin{equation}
  \label{eqn:cc}
  \mathbf{C} = f( \mathbf{I}, \mathbf{I'} ),
  \vspace{-2mm}
\end{equation}
we can get the cost volume $\mathbf{C} \in \mathbb{R}^{W \times H
\times L}$, which represents matching costs for each pixel at all
possible disparity levels. For a single pixel $i = (x_i, y_i)$,
where $x_i,y_i$ are pixel locations, its cost at disparity level $l$
can be denoted as a scalar, $\mathbf{C}(i,l)$. Various
methods can be used to compute the cost volume. For example, the
\emph{intensity + gradient} cost function
\cite{rhemann_11,Yang2012,XingMei} can be formulated as:
\begin{eqnarray}
  \label{eqn:cc_grd}
  \mathbf{C}(i,l) & = & (1-\alpha) \cdot \min( \| \mathbf{I}(i) - \mathbf{I'}(i_l) \|, \tau_1 ) \nonumber \\
  & & + \alpha \cdot \min( \| \nabla_x \mathbf{I}(i) - \nabla_x \mathbf{I'}(i_l) \|, \tau_2).
  \vspace{-2mm}
\end{eqnarray}
Here $\mathbf{I}(i)$ denotes the color vector of pixel $i$.
$\nabla_x$ is the grayscale gradient in $x$ direction. $i_l$ is the
corresponding pixel of $i$ with a disparity $l$,
i.e.\ $i_l = (x_i-l,y_i)$. $\alpha$ balances the color and gradient
terms and $\tau_1,\tau_2$ are truncation values.

% Para 3:
%   model cost aggregation

The cost volume $\mathbf{C}$ is typically very noisy
(Figure~\ref{fig:intro}). Inspired by the WLS formulation of the
denoising problem \cite{Milanfar2013}, the cost aggregation can be
formulated with the noisy input $\mathbf{C}$ as:
\begin{equation}
  \label{eqn:ca_opt}
   \mathbf{\tilde{C}}(i,l)
    = \mathop{\arg\min}_{z}{ \frac{1}{Z_i}\sum_{j \in N_i}{K(i,j) \| z - \mathbf{C}(j,l) \|^2} },
    \vspace{-2mm}
\end{equation}
where $N_i$ defines a neighboring system of $i$. $K(i,j)$ is the
similarity kernel \cite{Milanfar2013}, which measures the similarity between pixels $i$ and $j$, and $\mathbf{\tilde{C}}$ is the (denoised) cost volume.
$Z_i = \sum_{j \in N_i}{K(i,j)}$ is a normalization constant.
%It is obvious that cost aggregation can be viewed as denoising the cost volume.
The solution of this WLS problem is:

\begin{equation}
  \label{eqn:ca_solve}
  \mathbf{\tilde{C}}(i,l) = \frac{1}{Z_i} \sum_{ j \in N_i}{K(i,j)\mathbf{C}(j,l)}.
  \vspace{-2mm}
\end{equation}

% Para 4:
%   analysis different cost aggregation methods

Thus, like image filters \cite{Milanfar2013}, a cost
aggregation method corresponds to a particular instance of the
similarity kernel. For example, the \emph{BF} method \cite{Yoon2006}
adopted the spatial and photometric distances between two pixels to
measure the similarity, which is the same as the kernel function
used in the bilateral filter \cite{Tomasi1998}.
%One may note that the original cost aggregation method in
%\cite{Yoon2006} combined two similarity kernels defined in left and
%right images. However, as pointed out in \cite{Hosni2013}, using the
%asymmetric approach, i.e.\ only one similarity kernel from the left
%image, will have better performance when performing truncation of
%pixel-wise cost computation. Hence, we will adopt this asymmetric
%version in all the experiments.
Rhemann \etal \cite{rhemann_11} (\emph{GF}) adopted the kernel defined in the
guided filter \cite{hekaiming_10}, whose computational
complexity is independent of the kernel size. The \emph{NL}
method \cite{Yang2012} defines a kernel based on a geodesic
distance between two pixels in a tree structure. This approach was
further enhanced by making use of color segments, called a
segment-tree (\emph{ST}) approach \cite{XingMei}. A major difference
between filter-based \cite{Yoon2006,rhemann_11} and tree-based
\cite{Yang2012,XingMei} aggregation approaches is the action scope
of the similarity kernel, i.e.\ $N_i$ in Equation~\eqref{eqn:ca_solve}.
In filter-based methods, $N_i$ is a local window centered at $i$, but
in tree-based methods, $N_i$ is a whole image.
Figure~\ref{fig:intro} visualizes the effect of different action scope.
The filter-based methods hold some local similarity after the
cost aggregation, while tree-based methods tend to produce hard
edges between different regions in the cost volume.

Having shown that representative cost aggregation methods can be
formulated within a unified framework, let us recheck the cost
volume slices in Figure~\ref{fig:intro}. The slice, coming from
\emph{Teddy} stereo pair in the Middlebury dataset
\cite{middlebury}, consists of three typical scenarios: low-texture,
high-texture and near textureless regions (from left to right). The
four state-of-the-art cost aggregation methods all perform very well
in the high-texture area, but most of them fail in either  low-texture or near textureless region. %As above mentioned, human beings process stereoscopic correspondence across multiple scales.
For yielding highly accurate correspondence in those low-texture and
near textureless regions, the correspondence search should be
performed at the coarse scale \cite{Menz2003}. However, under the
formulation of Equation~\eqref{eqn:ca_opt}, costs are always aggregated 
at the finest scale, making it impossible to adaptively utilize
information from multiple scales. Hence, we need to reformulate the
WLS optimization objective from the scale space perspective.

% section tso (end)

\vspace{-1mm}
\section{Cross-Scale Cost Aggregation Framework} % (fold)
\label{sec:isr}
\vspace{-1mm}

% Para 1 :
%   briefly introduce SS

%In this section, we show that directly using Equation~(\ref{eqn:ca_opt})
%to tackle multi-scale cost volumes is equivalent to performing cost
%aggregation at each scale separately. Hence, we need to reformulate
%the optimization objective to enforce multi-scale interaction of
%cost volumes.

% Para 2:
%   add regularizer -> new model

It is straightforward to show that directly using
Equation~\eqref{eqn:ca_opt} to tackle multi-scale cost volumes is
equivalent to performing cost aggregation at each scale separately.
Firstly, we add a superscript $s$ to $\mathbf{C}$, denoting cost
volumes at different scales of a stereo pair, as $\mathbf{C}^s$,
where $s \in \{0, 1, \ldots, S\}$ is the scale parameter.
$\mathbf{C}^0$ represents the cost volume at the finest scale. The
multi-scale cost volume $\mathbf{C}^s$ is computed using the
downsampled images with a factor of $\eta^s$. Note that this approach
also reduces the search range of the disparity. The multi-scale
version of Equation~\eqref{eqn:ca_opt} can be easily expressed as:
\begin{equation}
  \label{eqn:ss_opt}
  \mathbf{\tilde{v}} = \mathop{\arg\min}_{\{z^s\}_{s=0}^{S}}{\sum_{s=0}^{S}{\frac{1}{Z^s_{i^s}}\!\sum_{j^s \in N_{i^s}}{K(i^s,j^s) \| z^s - \mathbf{C}^s(j^s,l^s) \|^2} }}.
  \vspace{-2mm}
\end{equation}

\noindent Here, $Z^s_{i^s} = \sum_{j^s \in N_{i^s}}{K(i^s,j^s)}$ is
a normalization constant. $\{i^s\}_{s=0}^{S}$ and
$\{l^s\}_{s=0}^{S}$ denote a sequence of corresponding variables at
each scale (Figure~\ref{fig:csca}), i.e.\ $i^{s+1}=i^s/\eta$
and $l^{s+1}=l^s/\eta$. %We set $\alpha$ to $2$ for simplicity, but a different choice is also possible.
$N_{i^s}$ is a set of neighboring pixels on the $s^{th}$ scale. In
our work, the size of $N_{i^s}$ remains the same for all scales,
meaning that more amount of smoothing is enforced on the coarser
scale. We use the vector $\mathbf{\tilde{v}}\!=\![\mathbf{\tilde{C}}^0(i^0,l^0), \mathbf{\tilde{C}}^1(i^1,l^1), \cdots, \mathbf{\tilde{C}}^S(i^S,l^S)]^T$ with $S+1$ components to denote the
aggregated cost at each scale. The solution of
Equation~\eqref{eqn:ss_opt} is obtained by performing cost aggregation
at each scale independently as follows:
\begin{equation}
  \label{eqn:ss_solve}
  \forall s, \mathbf{\tilde{C}}^s(i^s,l^s) = \frac{1}{Z^s_{i^s}} \sum_{ j^s \in N_{i^s}}{K(i^s,j^s)\mathbf{C}^s(j^s,l^s)}.
  \vspace{-2mm}
\end{equation}

Previous CTF approaches typically reduce the disparity
search space at the current scale by using a disparity map estimated from the cost volume at the coarser scale, often provoking the loss
of small disparity details. Alternatively, we directly enforce the
inter-scale consistency on the cost volume by adding a Generalized
Tikhonov regularizer into Equation~\eqref{eqn:ss_opt}, leading to the
following optimization objective:
\begin{eqnarray}
  \label{eqn:ss_reg_opt}
  \mathbf{\hat{v}}
  \! & = & \!\mathop{\arg\min}_{\{z^s\}_{s=0}^{S}}{
    (
      \sum_{s\!=\!0}^{S}{\!\frac{1}{Z^s_{i^s}}\!\sum_{j^s \in N_{i^s}}{
        \!K(i^s,j^s)
        \!\|\!z^s\!-\!\mathbf{C}^s(j^s,l^s)\!\|^2
        }
      }
    }
    \nonumber \\
    & & +
    \lambda \sum_{s=1}^{S}{
        \|z^s-z^{s-1}\|^2
      }
   ),
   \vspace{-2mm}
\end{eqnarray}
where $\lambda$ is a constant parameter to control the
strength of regularization. Besides, similar with $\mathbf{\tilde{v}}$, the vector $\mathbf{\hat{v}}\!=\![\mathbf{\hat{C}}^0(i^0,l^0), \mathbf{\hat{C}}^1(i^1,l^1), \cdots, \mathbf{\hat{C}}^S(i^S,l^S)]^T$ also has $S+1$ components to denote the cost at each scale.
The above optimization problem is convex. Hence, we can get the
solution by finding the stationary point of the optimization
objective. Let $F(\{z^s\}_{s=0}^{S})$ represent the optimization
objective in Equation~\eqref{eqn:ss_reg_opt}. For $s \in \{1, 2, \ldots,
S-1\}$, the partial derivative of $F$ with respect to $z^s$ is:
\begin{eqnarray}
  \label{eqn:ss_part}
  \frac{\partial F}{\partial z^s}\!&\!=\!&\!\frac{2}{Z^s_{i^s}}\!\sum_{j^s \in N_{i^s}}{\!K(i^s,j^s)(z^s-\mathbf{C}^s(j^s,l^s))}
  \nonumber \\
   & & + 2\lambda(z^s-z^{s-1}) - 2\lambda(z^{s+1}-z^s)
  \nonumber \\
  \!&\!=\!&\!2(\!-\!\lambda\!z^{s-1}\!+\!( 1\!+\!2\lambda ) z^s\!-\!\lambda\!z^{s+1}\!-\!\mathbf{\tilde{C}}^s(i^s,l^s)).
  \vspace{-2mm}
\end{eqnarray}
Setting $\frac{\partial F}{\partial z^s} = 0$, we get:
\begin{equation}
  \label{eqn:ss_eqn}
  -\lambda z^{s-1} + ( 1 + 2\lambda ) z^s - \lambda z^{s+1} = \mathbf{\tilde{C}}^s(i^s,l^s).
  \vspace{-2mm}
\end{equation}

\noindent It is easy to get similar equations for $s=0$ and $s=S$
. Thus, we have $S+1$ linear
equations in total, which can be expressed concisely as:
\begin{equation}
  \label{eqn:linear_sys}
  A \mathbf{\hat{v}} = \mathbf{\tilde{v}}.
\end{equation}

\noindent The matrix $A$ is an $(S+1) \times (S+1)$ tridiagonal constant
matrix, which can be easily derived from Equation~\eqref{eqn:ss_eqn}.
Since $A$ is tridiagonal, its inverse always exists. Thus,
\begin{equation}
  \label{eqn:final_eqn}
  \mathbf{\hat{v}} = A^{-1}\mathbf{\tilde{v}}.
  \vspace{-2mm}
\end{equation}
The final cost volume is obtained through the adaptive combination of the results of cost aggregation performed at different scales. Such adaptive combination enables the multi-scale interaction of the cost aggregation in the context of optimization.

\begin{figure}
  \begin{center}
    \includegraphics[width=0.4\textwidth]{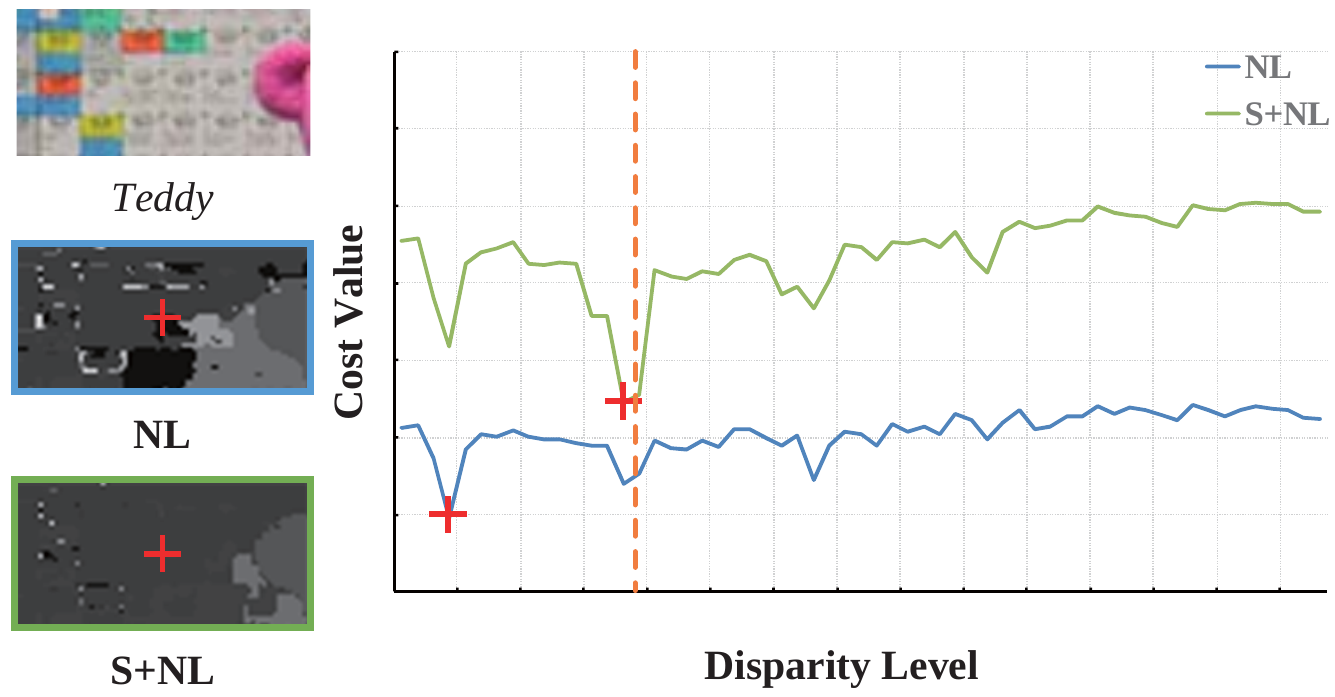}
  \end{center}
  \vspace{-2mm}
  \caption{The effect of inter-scale regularization. On the right side, we visualize two cost vectors of a single pixel (pixel location $(295,49)$) of \emph{Teddy} stereo pair. The blue line denotes the cost vector computed by \emph{NL} \cite{Yang2012} method. The green line is the cost vector after applying cross-scale cost aggregation (\emph{S+NL}). The red cross represents the minimal cost location for each cost vector and the vertical dash line denotes the ground truth disparity. On the left side, image and disparity patches centering on this pixel are shown. (Best viewed in color.)}
  \vspace{-4mm}
  \label{fig:cost}
\end{figure}

Finally, we use an example to show the effect of inter-scale regularization in Figure~\ref{fig:cost}. In this example, without cross-scale cost aggregation, there are similar local minima in the cost vector, yielding erroneous disparity. Information from the finest scale is not enough but when inter-scale regularization is adopted, useful information from coarse scales reshapes the cost vector, generating disparity closer to the ground truth.

% Then, assuming that we have got a robust cost volume $\mathbf{C}^{s-1}$ for scale $s-1$, how can we integrate this prior into the optimization objective (\ref{eqn:ca_opt}) to get a robust cost volume $\mathbf{C}^s$ for scale $s$? A straightforward solution is to add this prior as a Generalized Tikhonov regularizer:
% \begin{eqnarray}
%   \label{eqn:ca_ss}
%        \mathbf{\hat{c}}_{i}^{s} & = &  \mathop{\arg\min}_{\mathbf{x}} ( \frac{\alpha}{Z} \sum_{j \in N_i}{k_{ij} \| x - \mathbf{c}_{j}^{s} \|_2^2} \nonumber \\  & & + ( 1 - \alpha ) \| \mathbf{x} - \phi(\mathbf{c}_{i}^{s-1}) \|_2^2 ).
% \end{eqnarray}
% Here $ Z = \sum_{ j \in N_i }{k_{ij}} $ is the normlization constant, $\alpha$ is a parameter to control the effect of regularization and $\phi(\cdot)$ is a up-scaling operator to rescale $\mathbf{c}_{i}^{s-1}$. Solving this convex optimization problem, we can get the closed-form solution:
% \begin{equation}
%   \label{eqn:ca_solve}
%   \mathbf{\hat{c}}_{i}^{s} = \frac{\alpha}{Z} \sum_{j \in N_i}{k_{ij}\mathbf{c}_j^s} + (1-\alpha)\phi(\mathbf{c}_{i}^{s-1}).
% \end{equation}
% Indeed, the first part of this equationthe is the closed-form solution for the WLS problem defined in Equation~(\ref{eqn:ca_opt}). One may note that Equation~(\ref{eqn:ca_opt}) is a recursive definition. Thus we will show that by pre-computing cost volume at each scale, we can use Equation~(\ref{eqn:ca_opt}) iteratively to get the final cost volume.

% section isr (end)
\vspace{-1mm}
\section{Implementation and Complexity} % (fold)
\label{sec:impl}
\vspace{-1mm}
% Para 1:
%   introduce cross-scale theory
To build cost volumes for different scales (Figure~\ref{fig:csca}),
we need to extract stereo image pairs at different scales. In our
implementation, we choose the Gaussian Pyramid \cite{Burt1981},
which is a classical representation in the scale space theory. %\cite{Lindeberg1994a}.
The Gaussian Pyramid is obtained by successive smoothing and
subsampling ($\eta=2$). One advantage of this representation is that the
image size decreases exponentially as the scale level increases,
which reduces the computational cost of cost aggregation
on the coarser scale exponentially.
%Besides, it is straightforward to adopt other kinds of multi-scale
%representations \cite{Burt1981,Lindeberg1994a} into our framework.

\begin{algorithm}[htb]
    \centering
    \algblockdefx[INOUT]{Input}{Output}
    [1]{\textbf{Input:} #1}
    [1]{\textbf{Output:} #1}
    \begin{algorithmic}
        \Input{Stereo Color Image $\mathbf{I}$, $\mathbf{I'}$.}
        \begin{enumerate}
          \item Build Gaussian Pyramid $\mathbf{I}^s$, $\mathbf{I'}^s$, $s \in \{0, 1, \ldots, S\}$.
          \item Generate initial cost volume $\mathbf{C}^s$ for each scale by cost computation according to Equation~\eqref{eqn:cc}.
          \item Aggregate costs at each scale separately according to Equation~\eqref{eqn:ss_solve} to get cost volume $\mathbf{\tilde{C}}^s$.
          \item Aggregate costs across multiple scales according to Equation~\eqref{eqn:final_eqn} to get final cost volume $\mathbf{\hat{C}}^s$.
        \end{enumerate}
        \Output{Robust cost volume: $\mathbf{\hat{C}}^0$.}
    \end{algorithmic}
    \caption{Cross-Scale Cost Aggregation}
    \label{alg:ss_ca}
\end{algorithm}

% Para 2:
%   algorithm implementation
The basic workflow of the cross-scale cost aggregation is shown in
Algorithm~\ref{alg:ss_ca}, where we can utilize any existing cost
aggregation method in Step 3.
%Since we need to perform cost aggregation at each scale, one may argue that cross-scale cost
%aggregation involves too much computational cost. However, we will prove that
The computational complexity of our algorithm just
increases by a small constant factor, compared to conventional cost
aggregation methods. Specifically, let us denote the computational
complexity for conventional cost aggregation methods as $O(mWHL)$,
where $m$ differs with different cost aggregation methods.
%For example, $m$ equals the size of $N_i$ in \emph{BF} \cite{Yoon2006} while
%approximately equals $1$ in \emph{GF} \cite{rhemann_11}, \emph{NL}
%\cite{Yang2012} and \emph{ST} \cite{XingMei}.
%Since we use the Gaussian Pyramid \cite{Burt1981} to generate
%multi-scale representation for stereo images,
The number of pixels and disparities at scale $s$ are $\left\lfloor
\frac{WH}{4^{s}} \right\rfloor$ and $\left\lfloor \frac{L}{2^{s}}
\right\rfloor$ respectively. Thus the computational complexity of
Step 3 increases at most by $\frac{1}{7}$, compared to conventional
 cost aggregation methods, as explained below:
\begin{equation}
  \label{eqn:compl}
  \sum_{s=0}^{S}{\!\left(\!m\left\lfloor\!\frac{WHL}{8^{s}}\!\right\rfloor\!\right)}\!\leq\!\lim_{S \rightarrow \infty }{\left(\sum_{s=0}^{S}{\frac{mWHL}{8^{s}}}\right)} = \frac{8}{7} mWHL.
\end{equation}
Step 4 involves the inversion of the matrix $A$ with a size of
$(S+1)\times(S+1)$, but $A$ is a spatially invariant matrix,
with each row consisting of at most three nonzero elements, and thus its inverse can be pre-computed. Also, in Equation~\eqref{eqn:final_eqn}, the cost volume on the finest scale, $\mathbf{\hat{C}}^0(i^0,l^0)$,
is used to yield a final disparity map, and thus we need to compute
only
\begin{equation}
\label{eqn:linear_sys_sub}
\mathbf{\hat{C}}^0(i^0,l^0)=\sum\limits_{s = 0}^S
{A^{-1}(0,s)\mathbf{\tilde{C}}^s(i^s,l^s)},
\end{equation}

\noindent not $\mathbf{\hat{v}} = A^{-1}\mathbf{\tilde{v}}$. This
cost aggregation across multiple scales requires only a small amount
of extra computational load. In the following section, we will
analyze the runtime efficiency of our method in more details.

% section impl (end)
\vspace{-1mm}
\section{Experimental Result and Analysis} % (fold)
\label{sec:exp}
\vspace{-1mm}

\begin{figure*}
  \begin{center}
  % Normal CA
  \subfigure[BOX ($14.23\%$)]{
    \includegraphics[width=0.18\textwidth]{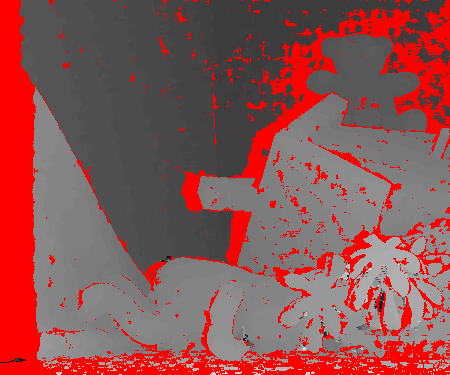}
  }
  \subfigure[NL ($8.60\%$)]{
    \includegraphics[width=0.18\textwidth]{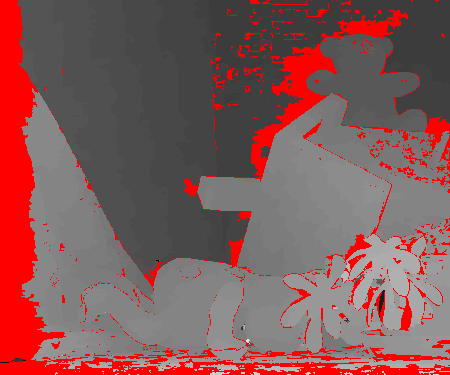}
  }
  \subfigure[ST ($9.78\%$)]{
    \includegraphics[width=0.18\textwidth]{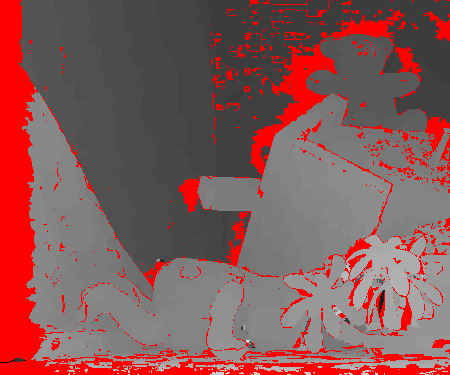}
  }
  \subfigure[BF ($10.24\%$)]{
    \includegraphics[width=0.18\textwidth]{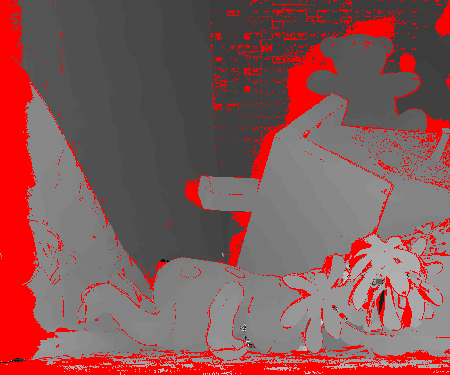}
  }
  \subfigure[GF ($8.25\%$)]{
    \includegraphics[width=0.18\textwidth]{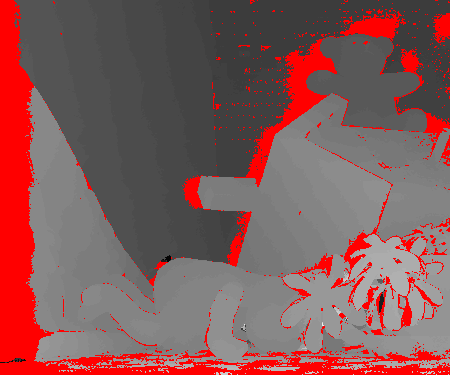}
  }
  \vspace{-2mm}
  % CSCA
  \subfigure[S+BOX ($11.18\%$)]{
    \includegraphics[width=0.18\textwidth]{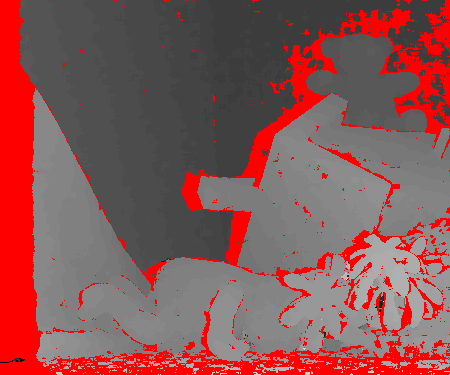}
  }
  \subfigure[S+NL ($5.74\%$)]{
    \includegraphics[width=0.18\textwidth]{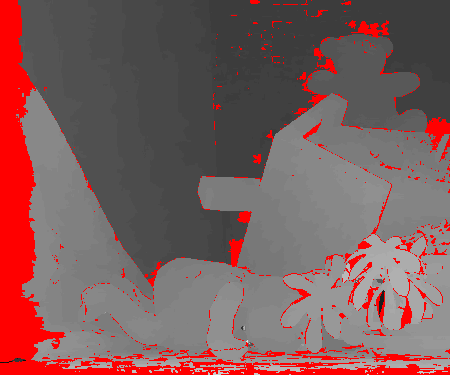}
  }
  \subfigure[S+ST ($6.22\%$)]{
    \includegraphics[width=0.18\textwidth]{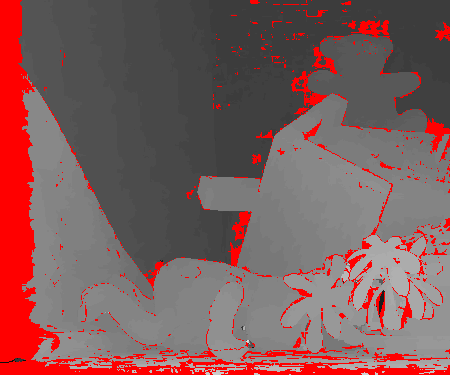}
  }
  \subfigure[S+BF ($8.17\%$)]{
    \includegraphics[width=0.18\textwidth]{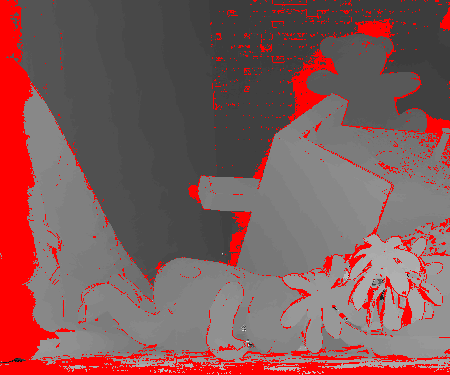}
  }
  \subfigure[S+GF ($6.99\%$)]{
    \includegraphics[width=0.18\textwidth]{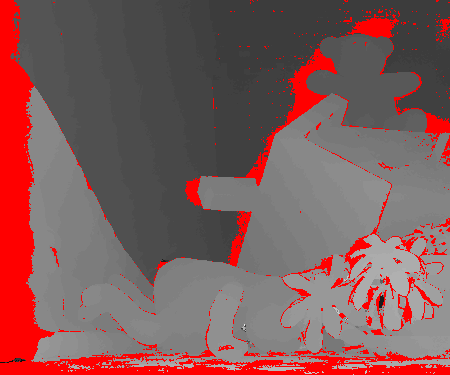}
  }
  \end{center}
  \vspace{-2mm}
  \caption{Disparity maps of \emph{Teddy} for all cost aggregation methods (with no disparity refinement techniques). The \emph{non-occ} error rate is shown in each subtitle. Red pixels indicate erroneous pixels, where the absolute disparity error is larger than $1$. (Best viewed in color.)} 
  \vspace{-5mm}
  \label{fig:mdb}
\end{figure*}

% Para 1:
%   use 2 datasets
%   parameter settings
In this section, we use Middlebury \cite{scharstein_02}, KITTI
\cite{Geiger2012} and New Tsukuba \cite{Peris} datasets to validate
that when integrating state-of-the-art cost aggregation methods,
such as \emph{BF} \cite{Yoon2006}, \emph{GF} \cite{rhemann_11},
\emph{NL} \cite{Yang2012} and \emph{ST} \cite{XingMei}, into our
framework, there will be significant performance improvements.
Furthermore, we also implement the simple box filter aggregation
method (named as \emph{BOX}, window size is $7 \times 7$) to serve
as a baseline, which also becomes very powerful when integrated into
our framework. For \emph{NL} and
\emph{ST}, we directly use the C++ codes provided by the
authors\footnote{{\footnotesize\url{http://www.cs.cityu.edu.hk/~qiyang/publications/cvpr-12/code/}}}\textsuperscript{,}\footnote{\url{http://xing-mei.net/resource/page/segment-tree.html}},
and thus all the parameter settings are identical to
those used in their implementations. For \emph{GF}, we implemented
our own C++ code by referring to the author-provided software
(implemented in
MATLAB\footnote{\url{https://www.ims.tuwien.ac.at/publications/tuw-202088}}) in order to process high-resolution images from KITTI and
New Tsukuba datasets efficiently. %In the supplementary material, we show that our implementation is almost the same as the author's.
For \emph{BF}, we implemeted the asymmetric version as suggested by \cite{Hosni2013}.
The local WTA strategy is adopted to generate a disparity map. In order to compare different cost aggregation methods fairly, no disparity refinement technique is employed, unless we explicitly declare. $S$ is set to 4, i.e.\ totally five scales are
used in our framework. For the regularization parameter $\lambda$,
we set it to $0.3$ for the Middlebury dataset, while setting it to
$1.0$ on the KITTI and New Tsukuba datasets for more regularization,
considering these two datasets contain a large portion of
textureless regions.

\subsection{Middlebury Dataset} % (fold)
\label{sub:middle}

% Para 1:
%   introduce middlebury dataset
%   introduce our evaluation metric

The Middlebury benchmark \cite{middlebury} is a de facto standard
for comparing existing stereo matching algorithms. In the benchmark
\cite{middlebury}, four stereo pairs (\emph{Tsukuba}, \emph{Venus},
\emph{Teddy}, \emph{Cones}) are used to rank more than $100$ stereo
matching algorithms. In our experiment, we adopt these four stereo
pairs. In addition, we use `Middlebury 2005' \cite{Scharstein2007}
($6$ stereo pairs) and `Middlebury 2006' \cite{hirschmuller_07}
($21$ stereo pairs) datasets, which involve more complex scenes.
Thus, we have $31$ stereo pairs in total, denoted as \emph{M31}. It is worth mentioning that during our experiments, all
local cost aggregation methods perform rather bad (error rate of
non-occlusion (\emph{non-occ}) area is more than $20\%$) in $4$
stereo pairs from Middlebury 2006 dataset, i.e.\ \emph{Midd1},
\emph{Midd2}, \emph{Monopoly} and \emph{Plastic}. A common property
of these $4$ stereo pairs is that they all contain large textureless
regions, making local stereo methods fragile. In order to alleviate
bias towards these four stereo pairs, we exclude them from
\emph{M31} to generate another collection of stereo pairs, which we call \emph{M27}. We make statistics on both \emph{M31} and \emph{M27} (Table~\ref{tab:mdb}). We adopt the \emph{intensity + gradient} cost
function in Equation~\eqref{eqn:cc_grd}, which is widely used in
state-of-the-art cost aggregation methods
\cite{rhemann_11,XingMei,Yang2012}.

In Table~\ref{tab:mdb}, we show the average error rates of
\emph{non-occ} region for different cost aggregation methods on both
\emph{M31} and \emph{M27} datasets. We use the prefix `\textbf{S+}'
to denote the integration of existing cost aggregation methods into
cross-scale cost aggregation framework. \textbf{Avg
Non-occ} is an average percentage of bad matching pixels in
\emph{non-occ} regions, where the absolute disparity error is larger
than $1$. The results are encouraging: all cost aggregation methods
see an improvement when using cross-scale cost aggregation, and even
the simple \emph{BOX} method becomes very powerful (comparable to state-of-the-art on \emph{M27}) when using cross-scale cost aggregation. Disparity maps
of \emph{Teddy} stereo pair for all these methods are shown in
Figure~\ref{fig:mdb}, while others are shown in the supplementary
material due to space limit.

Furthermore, to follow the standard evaluation metric of the
Middlebury benchmark \cite{middlebury}, we show each cost
aggregation method's rank on the website (as of October 2013) in
Table~\ref{tab:mdb}. \textbf{Avg Rank} and \textbf{Avg Err} indicate
the average rank and error rate measured using \emph{Tsukuba},
\emph{Venus}, \emph{Teddy} and \emph{Cones} images \cite{middlebury}.
Here each method is combined with the state-of-the-art disparity
refinement technique from \cite{Yang2012} (For ST \cite{XingMei}, we list its original rank reported in the Middlebury benchmark \cite{middlebury}, since the same results was not reproduced using the author's C++ code). The rank also validates the effectiveness of our framework. We also reported the running time for \emph{Tsukuba} stereo pair on a PC with a 2.83~GHz CPU and 8 GB of memory. As mentioned before, the computational overhead is
relatively small. To be specific, it consists of the cost
aggregation of $\mathbf{\tilde{C}}^s$ ($s\in \{0, 1, \cdots,S\}$) and
the computation of Equation~\eqref{eqn:linear_sys_sub}.

% Table
%   show middlebury results
\begin{table}
  \small
  \begin{center}
    \begin{tabular}{l|cc|cc|c}
      \hline
      \multirow{2}{*}{{Method}} & \multicolumn{2}{c|}{{Avg Non-occ}(\%)} & {Avg} & {Avg} & {Time}
      \\
      & \emph{M31} & \emph{M27} & {Rank} & {Err}(\%) & (s)\\
      \hline
      BOX & 15.45 & 10.7 & 59.6 &  6.2 & 0.11 \\
      {S+BOX} & \textbf{13.09} & \textbf{8.55} & \textbf{51.9} & \textbf{5.93} & {0.15} \\
      \hline
      NL\cite{Yang2012}     & 12.22 & 9.44 & 41.2 & 5.48 & 0.29  \\
      {S+NL}  & \textbf{11.49} & \textbf{8.73} & \textbf{39.4} & \textbf{5.2} & {0.37} \\
      \hline
      ST\cite{XingMei}     & 11.52 & 8.95  & 31.6 & 5.35 & 0.2  \\
      {S+ST} & \textbf{10.51} & \textbf{8.07}  & \textbf{27.9} & \textbf{4.97} & {0.29} \\
      \hline
      BF\cite{Yoon2006}     & 12.26 & 8.77  & 48.1 & 5.89 & 60.53 \\
      {S+BF} & \textbf{10.95} & \textbf{8.04}  & \textbf{40.7} & \textbf{5.56} & {70.62} \\
      \hline
      GF\cite{rhemann_11}     & 10.5 & 6.84  & 40.5 & 5.64 & 1.16 \\
      {S+GF} & \textbf{9.39} & \textbf{6.20}  & \textbf{37.7} & \textbf{5.51} & {1.32} \\
      \hline
    \end{tabular}
  \end{center}
  \caption{Quantitative evaluation of cost aggregation methods on the
  Middlebury dataset. The prefix {`S+'} denotes our cross-scale
  cost aggregation framework. For the rank part (column $4$ and $5$), the disparity results were refined with the same disparity refinement technique \cite{Yang2012}. }
  %\emph{M31} = Middlebury 2005
  %\cite{Scharstein2007} + 2006 \cite{hirschmuller_07} datasets and
  %\emph{M27}=\emph{M31}-{`Midd1', `Midd2', `Monopoly', `Plastic'}.
  %\textbf{Time} represents the running time for \emph{Tsukuba} stereo pair.
  \vspace{-2mm}
  \label{tab:mdb}
\end{table}

% subsection middle (end)

\subsection{KITTI Dataset} % (fold)
\label{sub:kitti_dataset}

% Table
%   GRD for KITTI
% \begin{table}[tb]
%   \begin{center}
%     \begin{tabular}{ l | c | c | c | c}
%       \hline
%       {\bf Method} & {\bf Out-Noc} & {\bf Out-All} & {\bf Avg-Noc} & {\bf Avg-All}\\
%       \hline
%       BOX & 54.21 \% & 55.24 \% &  29.35 px & 29.69 px \\
%       NL\cite{Yang2012} & 39.72 \% & 40.99 \% & 13.98 px & 14.79 px \\
%       ST\cite{XingMei} & 40.39 \% & 41.68 \% & 15.41 px & 16.18 px \\
%       % BF\cite{Yoon2006} & 00.00 \% & 00.00 \% & 00.00 px & 00.00 px \\
%       GF\cite{rhemann_11} & 36.25 \% & 37.67 \% & 19.50 px & 20.14 px \\
%       \hline
%     \end{tabular}
%   \end{center}
%   \caption{Quantitative comparison of cost aggregation methods on KITTI dataset when using \emph{intensity + gradient} as cost function. \textbf{Out-Noc}: percentage of erroneous pixels in non-occluded areas; \textbf{Out-All}: percentage of erroneous pixels in total; \textbf{Avg-Noc}: average disparity error in non-occluded areas; \textbf{Avg-All}: average disparity error in total.}
%   \vspace{-5mm}
%   \label{tab:grd_kitti}
% \end{table}

% Table
%   CEN for KITTI
\begin{table}
  \small
  \begin{center}
    \begin{tabular}{ l | c | c | c | c}
      \hline
      {Method} & {Out-Noc} & {Out-All} & {Avg-Noc} & {Avg-All}\\
      \hline
      BOX             & 22.51 \% & 24.28 \% & 12.18 px & 12.95 px \\
      {S+BOX} & \textbf{12.06} \% & \textbf{14.07} \% & \textbf{3.54} px & \textbf{4.57} px \\
      \hline
      NL\cite{Yang2012} & \textbf{24.69} \% & \textbf{26.38} \% & 4.36 px & 5.54 px \\
      {S+NL}  & {25.41} \% & {27.08} \% & \textbf{4.00} px & \textbf{5.20} px \\
      \hline
      ST\cite{XingMei} & \textbf{24.09} \% & \textbf{25.81} \% & 4.31 px & 5.47 px \\
      {S+ST}  & {24.51} \% & {26.22} \% & \textbf{3.82} px & \textbf{5.02} px \\
      \hline
      % BF\cite{Yoon2006} & 00.0 \% & 00.0 \% & 00.0 px & 00.0 px \\
      % \textbf{S+BF} & \textbf{00.0} \% & \textbf{00.0} \% & \textbf{00.0} px & \textbf{00.0} px \\
      % \hline
      GF\cite{rhemann_11} & 12.50 \% & 14.51 \% & 4.64 px & 5.69 px \\
      {S+GF} & \textbf{9.66} \% & \textbf{11.73} \% & \textbf{2.19} px & \textbf{3.36} px \\
      \hline
    \end{tabular}
  \end{center}
  \caption{Quantitative comparison of cost aggregation methods on KITTI dataset. \textbf{Out-Noc}: percentage of erroneous pixels in non-occluded areas; \textbf{Out-All}: percentage of erroneous pixels in total; \textbf{Avg-Noc}: average disparity error in non-occluded areas; \textbf{Avg-All}: average disparity error in total.}
  \vspace{-3mm}
  \label{tab:cen_kitti}
\end{table}

% Para 1:
%   introduce KITTI dataset

The KITTI dataset \cite{Geiger2012} contains $194$ training image
pairs and $195$ test image pairs for evaluating stereo matching
algorithms. For the KITTI dataset, image pairs are captured under real-world illumination condition and almost all image pairs have a large portion of textureless regions, \eg walls and roads \cite{Geiger2012}. During our experiment, we use the whole $194$ training image pairs with ground truth disparity maps available. The evaluation metric is the same as the KITTI benchmark \cite{kitti} with an error threshold $3$. Besides, since \emph{BF} is too slow for high resolution images (requiring more than one hour to process one stereo pair), we omit \emph{BF} from evaluation.

% Para 2:
%   effect of cost function

Considering the illumination variation on the KITTI dataset, we
adopt \emph{Census Transform} \cite{zabih_94}, which is proved to be
powerful for robust optical flow computation \cite{hafner13}. We
show the performance of different methods when integrated into
cross-scale cost aggregation in Table~\ref{tab:cen_kitti}. Some
interesting points are worth noting. Firstly, for \emph{BOX} and
\emph{GF}, there are significant improvements when using cross-scale
cost aggregation. Again, like the Middlebury dataset, the simple
\emph{BOX} method becomes very powerful by using cross-scale cost
aggregation. However, for \emph{S+NL} and \emph{S+ST}, their
performances are almost the same as those without cross-scale cost
aggregation, which are even worse than that of \emph{S+BOX}. This may
be due to the non-local property of tree-based cost aggregation
methods. For textureless slant planes, \eg~roads, tree-based
methods tend to overuse the \emph{piecewise constancy} assumption
and may generate erroneous fronto-parallel planes. Thus, even though the cross-scale cost aggregation is adopted, errors in textureless slant planes are not fully addressed. Disparity maps for
all methods are presented in the supplementary material, which also
validate our analysis.

\subsection{New Tsukuba Dataset} % (fold)
\label{sub:new_tsuku}

% Table
%   GRD for New Tsukuba
\begin{table}
  \small
  \begin{center}
    \begin{tabular}{ l | c | c | c | c}
      \hline
      {Method} & {Out-Noc} & {Out-All} & {Avg-Noc} & {Avg-All}\\
      \hline
      BOX             & 31.08 \% & 37.70 \% & 7.37 px & 10.72 px \\
      {S+BOX} & \textbf{18.82} \% & \textbf{26.50} \% & \textbf{3.92} px & \textbf{7.44} px \\
      \hline
      NL\cite{Yang2012} & 21.88 \% & 26.72 \% & 4.12 px & 6.40 px \\
      {S+NL}  & \textbf{19.84} \% & \textbf{24.50} \% & \textbf{3.65} px & \textbf{5.73} px \\
      \hline
      ST\cite{XingMei} & 21.68 \% & 27.07 \% & 4.33 px & 7.02 px \\
      {S+ST}  & \textbf{18.99} \% & \textbf{24.16} \% & \textbf{3.60} px & \textbf{5.96} px \\
      \hline
      % BF\cite{Yoon2006} & 00.0 \% & 00.0 \% & 00.0 px & 00.0 px \\
      % \textbf{S+BF} & \textbf{00.0} \% & \textbf{00.0} \% & \textbf{00.0} px & \textbf{00.0} px \\
      % \hline
      GF\cite{rhemann_11} & 23.42 \% & 30.34 \% & 6.35 px & 9.86 px \\
      {S+GF} & \textbf{14.40} \% & \textbf{21.78} \% & \textbf{3.10} px & \textbf{6.38} px \\
      \hline
    \end{tabular}
  \end{center}
  \caption{Quantitative comparison of cost aggregation methods on New Tsukuba dataset.}
  \vspace{-3mm}
  \label{tab:grd_newT}
\end{table}

The New Tsukuba Dataset \cite{Peris} contains $1800$ stereo pairs
with ground truth disparity maps. These pairs consist of a one
minute photorealistic stereo video, generated by moving a stereo
camera in a computer generated 3D scene. Besides, there are $4$
different illumination conditions: \emph{Daylight},
\emph{Fluorescent}, \emph{Lamps} and \emph{Flashlight}. In our
experiments, we use the \emph{Daylight} scene, which has a
challenging real world illumination condition \cite{Peris}. Since
neighboring frames usually share similar scenes, we sample the
$1800$ frames every second to get a subset of $60$ stereo pairs,
which saves the evaluation time. We test both \emph{intensity +
gradient} and \emph{Census Transform} cost functions, and
\emph{intensity + gradient} cost function gives better results in
this dataset. Disparity level of this dataset is the same as the
KITTI dataset, i.e.\ $256$ disparity levels, making BF
\cite{Yoon2006} too slow, so we omit BF from evaluation.

Table~\ref{tab:grd_newT} shows evaluation results for different cost
aggregation methods on New Tsukuba dataset. We use the same
evaluation metric as the KITTI benchmark \cite{kitti} (error
threshold is $3$). Again, all cost aggregation methods see an
improvement when using cross-scale cost aggregation.

% subsection new_tusk (end)

\subsection{Regularization Parameter Study} % (fold)
\label{sub:regu}

\begin{figure}
  \begin{center}
    \includegraphics[width=0.42\textwidth]{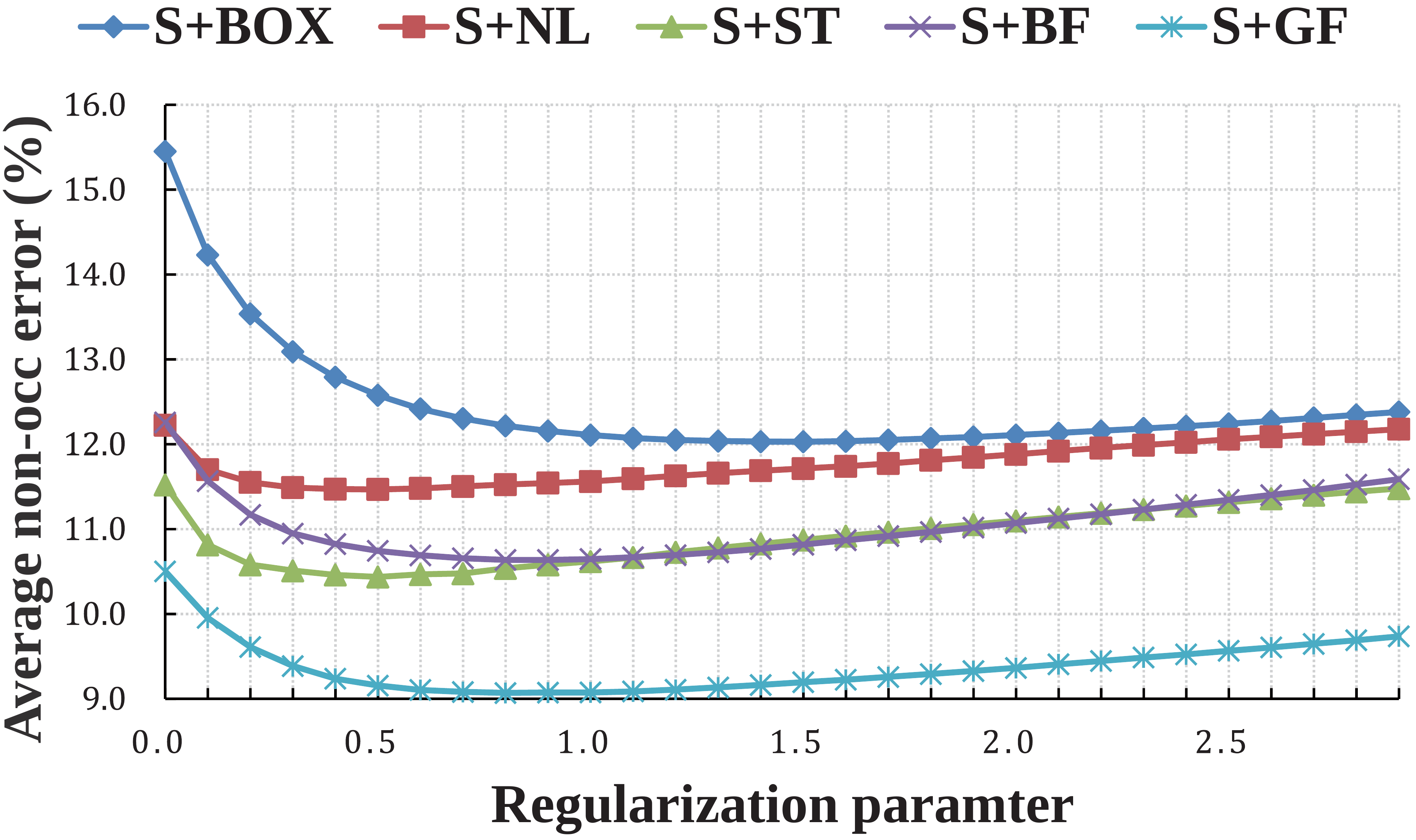}
  \end{center}
  \vspace{-3mm}
  \caption{The effect of varying inter-scale regularization parameter for different methods.}
  \vspace{-4mm}
  \label{fig:paramter}
\end{figure}

% % Table
% %   disparity refinement
% \begin{table}
%   \begin{center}
%     \begin{tabular}{l|cc}
%       \hline
%       & \multicolumn{2}{c}{\textbf{Avg Non-Occ Err($\%$)}}
%       \\
%       \rb{\textbf{Method}}
%       &\footnotesize{\textit{w/o DR}}&\footnotesize{\textit{w/ DR}}
%       \\
%       \hline
%       BOX(\emph{baseline}) & 15.45 & 12.32 \\
%       \textbf{S+BOX} & \textbf{12.11} & \textbf{10.70} \\
%       \hline
%       NL\cite{Yang2012}     & 12.22 & 11.79 \\
%       \textbf{S+NL}  & \textbf{11.56} & \textbf{11.62} \\
%       \hline
%       ST\cite{XingMei}     & 11.52 & 10.38 \\
%       \textbf{S+ST} & \textbf{10.62} & \textbf{10.20} \\
%       \hline
%       BF\cite{Yoon2006}     & 12.26 & 10.93 \\
%       \textbf{S+BF} & \textbf{10.65} & \textbf{10.07} \\
%       \hline
%       GF\cite{rhemann_11}     & 10.5 & 9.82 \\
%       \textbf{S+GF} & \textbf{9.08} & \textbf{8.87} \\
%       \hline
%     \end{tabular}
%   \end{center}
%   \caption{The effect of disparity refinement for different methods. All methods are evaluated on \emph{Middlebury31} and \emph{DR} is short for \emph{disparity refinement}.}
%   \vspace{-5mm}
%   \label{tab:dr}
% \end{table}

% Para 1:
%   effect of inter-scale regularization parameter

The key parameter in Equation~\eqref{eqn:ss_reg_opt} is the
regularization parameter $\lambda$. By adjusting this parameter, we
can control the strength of inter-scale regularization as shown in
Figure~\ref{fig:paramter}. The error rate is evaluated on \emph{M31}. When $\lambda$ is set to $0$, inter-scale
regularization is prohibited, which is equivalent to performing cost
aggregation at the finest scale. When regularization is introduced,
there are improvements for all methods. As $\lambda$ becomes large, the regularization term dominates the optimization, causing the cost volume of each scale to be purely
identical. As a result, fine details of disparity maps are missing
and error rate increases. One may note that it will generate better results by choosing different $\lambda$ for different cost
aggregation methods, though we use consistent $\lambda$ for all methods.
% Furthermore, as we discussed in Section~\ref{sub:kitti_dataset}, tree-based cost aggregation is less sensitive to the inter-scale regularization.

% Para 2:
%   disparity refinement results

% From previous experiments, it is obvious that state-of-the-art cost aggregation methods can benefit cross-scale cost aggregation, which may have similar effect with \emph{disparity refinement} (Step $4$ in Figure~\ref{fig:proc}). Thus, it is assumed that combining cross-scale cost aggregation with disparity refinement will generate better disparity maps. We conduct experiments to validate this assumption. We directly use the disparity refinement code of \emph{GF} \cite{rhemann_11} and show error rates of different methods on \emph{Middlebury31} dataset in Table~\ref{tab:dr}. Almost all results are consistent with our assumption except for \emph{S+NL} where the best performance is achieved without disparity refinement. According to \cite{Yang2012}, different cost aggregation methods may have bias to certain disparity refinement methods, and the combination of \emph{NL} and \emph{GF}'s disparity refinement method leads to worse results \cite{Yang2012}.

% subsection regu (end)

% section exp (end)
\vspace{-1mm}
\section{Conclusions and Future Work} % (fold)
\label{sec:con}
\vspace{-1mm}
% Para 1:
%    conclusion

In this paper, we have proposed a cross-scale cost aggregation
framework for stereo matching. This paper is not intended to present
a completely new cost aggregation method that yields a highly
accurate disparity map. Rather, we investigate the scale space
behavior of various cost aggregation methods. Extensive experiments on three
datasets validated the effect of cross-scale cost aggregation.
Almost all methods saw improvements and even the simple box
filtering method combined with our framework achieved very good
performance.

% Para 2:
%   future work
Recently, a new trend in stereo vision is to solve the
correspondence problem in continuous plane parameter space rather
than in discrete disparity label space
\cite{bleyer_11,Lu2013,Yamaguchi2013}. These methods can handle
slant planes very well and one probable future direction is to investigate
the scale space behavior of these methods.

% section con (end)
\vspace{-1mm}
\section{Acknowledgement} % (fold)
\label{sec:ack}
\vspace{-1mm}

% This work was supported by the National Basic Research
% Program (973 Program) of China under Grant
% 2012CB316301, and Basic Research Foundation of Tsinghua
% National Laboratory for Information Science and
% Technology (TNList). This work was also supported in part
% to Dr. Qi Tian by ARO grant W911NF-12-1-0057, NSF IIS
% 1052851, Faculty Research Awards by Google, NEC Laboratories
% of America and FXPAL, UTSA START-R award
% and NSFC 61128007, respectively. This work was also supported
% by the Singapore National Research Foundation under
% its International Research Centre at Singapore Funding
% Initiative and administered by the IDM Programme Office.

This work was supported by XXXXXXXXXXXX.

% section ack (end)
%
% bibliography
%
{\footnotesize
\bibliographystyle{myieee}
\bibliography{library}
}
\end{document}